# AdvJND: Generating Adversarial Examples with Just Noticeable Difference

Zifei Zhang, Kai Qiao, Lingyun Jiang, Linyuan Wang, Jian Chen, and Bin Yan

## Abstract

Compared with traditional machine learning models, deep neural networks perform better, especially in image classification tasks. However, they are vulnerable to adversarial examples. Adding small perturbations on examples causes a good-performance model to misclassify the crafted examples, without category differences in the human eyes, and fools deep models successfully. There are two requirements for generating adversarial examples: the attack success rate and image fidelity metrics. Generally, perturbations are increased to ensure the adversarial examples' high attack success rate; however, the adversarial examples obtained have poor concealment. To alleviate the tradeoff between the attack success rate and image fidelity, we propose a method named AdvJND, adding visual model coefficients, just noticeable difference coefficients, in the constraint of a distortion function when generating adversarial examples. In fact, the visual subjective feeling of the human eyes is added as a priori information, which decides the distribution of perturbations, to improve the image quality of adversarial examples. We tested our method on the FashionMNIST, CIFAR10, and MiniImageNet datasets. Our adversarial examples keep high image quality under slightly decreasing attack success rate. Since our AdvJND algorithm yield gradient distributions that are similar to those of the original inputs, the crafted noise can be hidden in the original inputs, improving the attack concealment significantly.

## 1. Introduction

Deep neural networks (DNNs) are effective for completing many important but difficult tasks like computer vision [1-4], nature language processing [5-8], etc., and can achieve state-of-the-art performances in these tasks. Furthermore, they have approached human levels of performance in some specific tasks. Thus, we can assume that artificial intelligence is moving toward human intelligence step by step. However, Szegedy made an intriguing discovery: DNNs are vulnerable to adversarial examples [9]; he first proposed the concept of adversarial examples in image classification. A good-performance DNN model misclassify inputs modified by adding small, imperceptible perturbations, which is hard to distinguish for humans. And adversarial examples are used to attack such applications like face recognition[10, 11], autonomous driving car [12, 13] and malware detection [14]. Obviously, adversarial examples are blind spots of deep models. The problem of generating adversarial examples can be regarded as an optimization problem, in which the target perturbations are minimized when the predicted label is not equal to the true label. The mathematical formula is decribed as follows:

$$\begin{aligned} & \min \ D(x, x+r), \\ & \text{s.t.} \ f(x+r) \neq f(x). \end{aligned} \quad (1)$$

Let $x$ be the input to the model, $r$ the perturbation, $D(x, x+r)$ the distortion function between adversarial examples and their original inputs, and $f(x)$ the predicted label of the model. As shown in formula (1), there are two requirements for generating adversarial examples. One is to generate a misclassified example to attack successfully, and the other is to generate the smallest possible distortion value. These requirements ensure that the adversarial examples are similar to the original inputs and that high image fidelity is guaranteed. Because of the security threat of DNNs, adversarial examples have garnered significant attention among researchers, especially in the security critical applications. Classic methods for generating adversarial examples on deep learning have been established. Based on the adversarial setting criteria to sort, white-box attack represents to directly acquire all information, like training datasets, model architecture and so on. However, black-box attack means to get information by querying model indirectly. And the proposed methods usually use the $L_p$ norm ($L_0$, $L_2$, $L_\infty$) to classify the adversarial examples, which is used for constraining the perturbations. That is, in the definition of the distortion function $D(x, x+r)$, the $L_p$ norm is used as a distance metric to measure the similarity between the adversarial examples and the original inputs. Typically, Jiawei Su et al. [15] proposed the one pixel attack method with the $L_0$ norm constraint, which changes by only one or several pixels [16, 17] in a picture but results in a significant changes compared with the original image of the poor attack concealment with obvious altered traces. Additionally, a lower attack success rate is resulted. Szegedy et al. proposed a method to generate adversarial examples with box-constrained L-BFGS [9] via back-propagation to obtain gradient information. Moosavi-Dezfooli et al. proposed a method to search the minimum perturbations to a classified boundary, named DeepFool [18], with the high images fidelity and attack success rates. Both of them take the $L_2$ norm constraint, which interferes the entire picture. Adversarial examples which satisfy the $L_2$ norm constraint are similar to the original inputs [18, 19]. However, it is time consuming to generate adversarial examples, which is inefficient. Goodfellow et al. proposed the fast gradient sign method (FGSM) [20] with the $L_\infty$ norm constraint, which fastly generates adversarial examples by maximizing the loss function, with low image fidelity and attack success rate. Furthermore, Kurakin Alexey et al. proposed an iterative fast gradient sign method (I-FGSM) [21] to improve FGSM.

We herein mainly discuss the $L_\infty$ norm constraint, restraining the maximum distance difference between the adversarial example and the original input. Generally, perturbations are increased to ensure the adversarial examples' high attack success rate; however, the adversarial examples obtained in this manner exhibit poor concealment. To alleviate the tradeoff between the attack success rate and image fidelity, we propose a method that adds visual model coefficients in the $L_\infty$ norm constraint. Because the $L_\infty$ norm constraint is an objective metric, the distribution of perturbation is disordered and some noisy pixels are sensitive to the human eyes. Sid Ahmed Fezza et al. [22] thought the $L_p$ norm did not correlate with human judgement and were not suitable as a distance metric. Adil Kaan Akan et al. [23] defined the machine's just noticeable difference with regularization terms, other than just noticeable difference of human visual perception. And they generated just noticeable difference adversarial examples, which attacked successfully just right. Different from that, we take the visual model coefficients into consideration, and think it can be added in the constraint to improve the images quality and guarantee high image fidelity. In fact, the visual subjective feeling of the human eyes is added as a priori information in the constraint to control the distribution of perturbations. In our study, we integrate the just noticeable difference (JND) coefficients into the $L_\infty$ norm constraint of the distortion function to complete above attentioned task.

JND coefficients are critical values at which a difference can be detected. Additionally, they reflect that the human eyes can recognize the threshold of an image change. In general, the JND model is applied in image encoding. There exists redundancy in images, which without de-redundancy would be transported with lower efficiency. And JND could determine the amount of tolerated distortions to guarantee the quality of the images. Image encoding with JND coefficients can improve coding efficiency significantly [24-26], called perceptual coding. In this study, we used the JND model of the image domain to hide noise. As shown in Figure 1, after adding Gaussian noise with a variance of 0.01 in the original input, the image is significantly interfered. When we constrain the noise with JND coefficients to control the distribution of noise, a human visual system (HVS) cannot distinguish the difference between the original input and the JND image, which proves the noise concealment ability of JND coefficients.

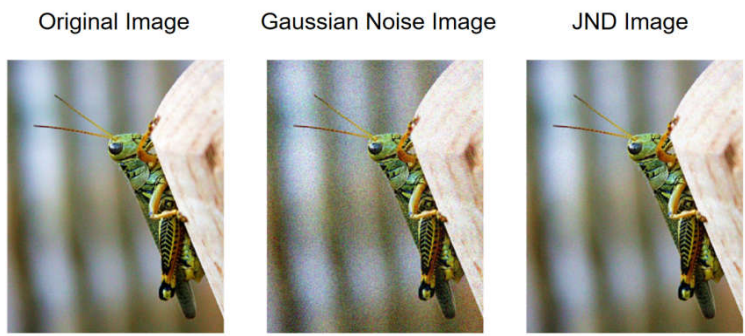

Figure 1. JND coefficients hide Gaussian noise. Left column: the original image. Middle column: the Gaussian noise image. Right column: the JND image.

JND coefficients can hide Gaussian noise because a region with large JND coefficients is a region with complex image textures. Additionally, it is difficult for our HVS to notice these changes in these regions, which are also called visual blind spots of the human eyes. The larger the JND coefficients, the higher are the thresholds, the greater is the redundancy, the smaller is the sensitivity of the human eyes, and the more noise can be embedded. Therefore, perturbations in regions with large JND coefficients are less likely to be detected. We integrate JND coefficients into the existing adversarial attack methods. Namely, we add JND coefficients to the $L_\infty$ norm constraint and define this method as AdvJND. The primarily contributions of this study are as follows:

(1) We suggest a method to integrate JND coefficients for generating adversarial examples. We add the visual subjective feeling of the human eyes as a priori information in the constraint to decide the distribution of perturbations and generate adversarial examples with gradients distribution similar to that of the original inputs. Hence, the crafted noise can be hidden in the original inputs, thus improving the attack concealment significantly.

(2) We demonstrate that generating adversarial examples with our algorithm costs less time than algorithms with the $L_2$ norm constraint like DeepFool, when the image quality and the attack success rate of their methods are approximate. Such fact proves that our AdvJND algorithm is more efficient.

In Section 2, we provide the implementation algorithm of AdvJND. The effects of AdvJND are shown in Section 3. In Section 4, we draw the conclusions.

## 2. Methodology

In our AdvJND algorithm, we should get some information in advance, like the original image's JND coefficients and the original perturbations from the target model's gradients. Hence, we compute the JND coefficients in Section 2.1, and adopt FGSM and I-FGSM methods to yield the original perturbations in Section 2.2. In Section 2.3, we introduce the complete AdvJND algorithm.

**2.1 JND Coefficients**

The JND coefficients are based on the representation of visual redundancy in psychology and physiology. The receiver of image information is the HVS. A JND spatial model in the image domain primarily includes two factors: luminance masking and texture masking. On one hand, according to the Weber's law, the luminance contrast of perception in HVS increases with the practical's luminance. On the other hand, since the complex texture area and excess noises are both high-frequency information, so that excess noises could be hided in the texture area easily. To better match the HVS characteristics, X. K. Yang [27] designed a nonlinear additive model for masking to give consideration to both luminance adaption and texture. And texture masking is determined by the average background luminance and the average luminance difference around a pixel [28, 29]. The JND coefficient of each pixel is obtained experimentally [27]. The formula is

$$jnd(i,j) = \max\left(f_1\left(bg(i,j), mg(i,j)\right), f_2(i,j)\right). \qquad (2)$$

Where $f_1(i,j)$ is the texture masking function, $f_2(i,j)$ is the luminance adaption function, $bg(i,j)$ and $mg(i,j)$ represent gradient changes of the average background luminance and neighboring points at point $(i,j)$, respectively.

Due to the visual redundancy in the image, there is a chance to embed noises in it. Furthermore, it is necessary for us to determine the magnitude of embedding noises to guarantee imperceptibility. Luckily, JND coefficients is related with HVS's sensitivity and helpful to embedding noises without perceptibility, which improves the attack concealment.

**2.2 Adversarial Attack Methods**

The paper is based on the white-box adversarial attack setting, instead of Curls & Whey [30], which concerntrates on improving adversarial image quality under the same query times in black-box setting.

In this section, we review the related studies of adversarial attack. We primarily introduce the FGSM and its extended algorithm I-FGSM and obtain the original perturbations from them. And our method performs improvements based on the FGSM and I-FGSM. The reason why we choose I-FGSM as a baseline is that I-FGSM is the state-of-the-art white-box attack based on $L_\infty$ norm constraint.

**FGSM.** The basic concept of the FGSM [20] is to optimize in the direction of increasing loss function, i.e., generating adversarial examples in the positive direction of the gradient. It exhibits two characteristics. One is that it generates adversarial examples fast, as it only performs one back-propagation without iteration. Another is that it measures the distance between the adversarial example and the original input using the $L_\infty$. These are the two main reasons for the obvious perturbations.

$$p = \varepsilon \cdot sign(\nabla_x J(\theta, x, y)). \tag{3}$$

$$x^{adv} = x + p \tag{4}$$

Where $\varepsilon$ represents the upper limit of perturbation, $\nabla_x J(\cdot)$ represents the gradient value of the loss function to the original input via back-propagation, $p$ represents the perturbation, $x$ represents the original input, and $x^{adv}$ represents the generated adversarial example.

**I-FGSM.** The I-FGSM [21] is the expansion of the FGSM, which computes perturbations iteratively instead of in a one-shot manner. Specifically, a $\varepsilon$ single value that changes in the direction of the gradient sign is replaced by a smaller $\alpha$ value; subsequently, the upper limit of the perturbation $\varepsilon$ is used as limiting the constraint.

$$x_0^{adv} = x. \tag{5}$$

$$Clip_{x,\varepsilon}\{x\} = \min(1, x+\varepsilon, \max(x-\varepsilon, x)). \tag{6}$$

$$x_{t+1}^{adv} = Clip_{x,\varepsilon}\{x_t^{adv} + \alpha \cdot sign(\nabla_x J(\theta, x, y))\}. \tag{7}$$

The I-FGSM achieves adversarial examples of better image quality than the one-shot FGSM. Meanwhile it implies more time costs.

**2.3 AdvJND Methods**

First, we are to calculate the JND coefficients of the original input and then normalize the processed JND coefficients to the $L_\infty$. Specifically, we normalize the original input pixels to [0,1], and calculate the JND coefficients on each channel independently to simplify the calculation. Although the JND coefficients can reflect the edge information to some extent, for a more obvious edge area and a better discrimination, we calculate the power values of the JND coefficients, which allow large values to become larger, and small values to become smaller, that is, values representing edge areas are dramatically larger than smooth areas. In this paper, we square the image's JND coefficients.

$$jnd_2 = jnd \times jnd. \tag{8}$$

On the other hand, after squaring, the obtained JND coefficients are close to the order of 1e-3. If perturbations added are directly controlled at 1e-3 or similar, it would be difficult to attack the image successfully although the perturbations obey the image's gradient distribution.

Thus, we discard the absolute values of the JND coefficients instead of their relative values, that is, we take JND coefficients to control the distribution of perturbations indirectly.

$$\lambda = \frac{p_{ori}}{\max(jnd_2)}. \tag{9}$$

$$k = \lambda \times jnd_2. \tag{10}$$

$p_{ori}$ represents the original perturbations from the FGSM or I-FGSM method, represents the scaled value, and k is the JND coefficients' relative values, which provide the critical information of the image texture location. Although the obtained adversarial examples are similar to the original inputs, their attack success rates are still lower than original adversarial examples'. In most cases, the large values of k primarily locate in the regions with complex textures, in which noise can be hided efficiently, and the small values of k locate in the smooth areas, in which our HVS are sensitive and easy to notice. Therefore, we decide the final values of k based on the location information. And our strategy is to reduce the small values of k in multiplies and calculate the final perturbations as follows.

$$\begin{cases} t = 1, \text{ if } k \geq \rho \\ t = \gamma, \text{ if } k < \rho \end{cases}. \tag{11}$$

$$p_{out} = k \times t. \tag{12}$$

We obtained the experience value experimentally. The threshold value $\rho = \frac{\varepsilon}{2}$, the reduced multiple $\gamma = \frac{1}{4}$, and $p_{out}$ represents the final adversarial perturbations. The AdvJND method is summarized in **Algorithm 1**.

**Algorithm 1** takes the FGSM method as an example to show the complete process of our AdvJND algorithm to generate adversarial examples. If we implement our AdvJND algorithm based on the I-FGSM method, take the output $x^{adv}$ as the input $x$, and repeat the procedures from step 3 to step 9 until satisfying the minimum condition or the maximum iterations.

**Algorithm 1** AdvJND: restrain JND coefficients to $L_\infty$ norm
---
**1** input: an image $x$, superior limit $\varepsilon$.
**2** output: an adversarial example $x^{adv}$.
**3** Computer JND coefficients of the image $x$
$$jnd(i,j) \leftarrow \max\left(f_1\left(bg(i,j), mg(i,j)\right), f_2(i,j)\right).$$
**4** Calculate the original perturbations $p_{ori}$
$$p \leftarrow \varepsilon \cdot sign\left(\nabla_x J(\theta, x, y)\right).$$
**5** Square JND coefficients
$$jnd_2 \leftarrow jnd \times jnd.$$
**6** Normalize JND coefficients to $L_\infty$ norm
$$k \leftarrow \frac{p_{ori}}{\max(jnd_2)} \times jnd_2.$$
**7** Set threshold
$$\begin{cases} t \leftarrow 1, \text{ if } k \geq \rho \\ t \leftarrow \gamma, \text{ if } k < \rho \end{cases}.$$
**8** Obtain Perturbations $p_{out}$
$$p_{out} \leftarrow k \times t.$$
**9** Get the final adversarial example
$$x^{adv} \leftarrow x + p_{out}$$
**10** return $x^{adv}$

## 3. Experiments

In this section, experiments on the FashionMNIST [31], CIFAR10 [32], and MiniImageNet datasets (using 1000 images from ILSVR2012 [33] test dataset, 1925 pictures in total, and the reason why we take the MiniImageNet dataset is that it can not guarantee the high recognition accuracy in classification tasks with the whole ImageNet dataset, and in order to show the effectiveness of our attack algorithm, we validate the MiniImageNet with high accuarcy.) are used to validate our AdvJND method, and these datasets correspond to network architectures LeNet-5 [34], VGG16 [35], and Inception_v3 [36], respectively. We demonstrate the advantages of the FGSM-JND and I-FGSM-JND algorithms over the original attack methods in Section 3.1. And the proposed AdvJND algorithm adopts a general approach of the $L_\infty$ constraint to generate adversarial examples. In Section 3.2, we compare the efficiency between the I-FGSM-JND and DeepFool algorithms.

### 3.1 AdvJND

The core of AdvJND is integrating JND coefficients into the $L_\infty$ constraint, i.e., using the visual subjective feeling of the human eyes as priori information to control the distribution of perturbations. More similar adversarial examples are generated though the attack success rate slightly decreases within an acceptable range.

### 3.1.1 FGSM vs. FGSM-JND

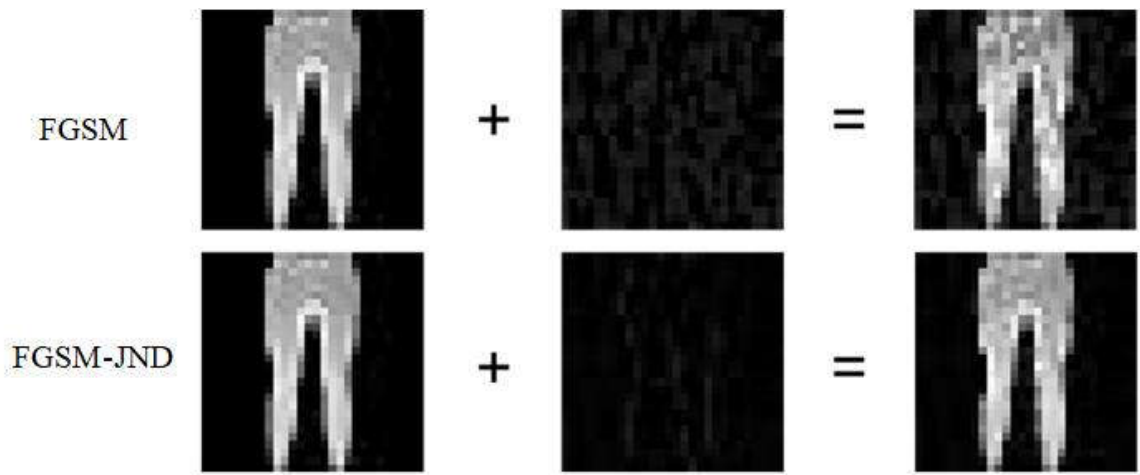

Figure 2. FGSM vs. FGSM-JND on the FashionMNIST dataset.

The FGSM-JND is obtained by integrating JND coefficients into the FGSM. As shown in Figure 2, the perturbations generated by the FGSM are distributed over the entire image, but the perturbations generated by the FGSM-JND are distributed over the edge region of the "pants". The adversarial examples generated by FGSM are rough and modified obviously, but the adversarial examples generated by our algorithm are smooth and more similar to the original inputs, since our FGSM-JND algorithm can effectively control perturbations in such smooth regions with the location of small JND coefficients and mainly hide noise in regions with the location of large JND coefficients to ensure its adversarial capacity.

**3.1.2 I-FGSM vs. I-FGSM-JND**

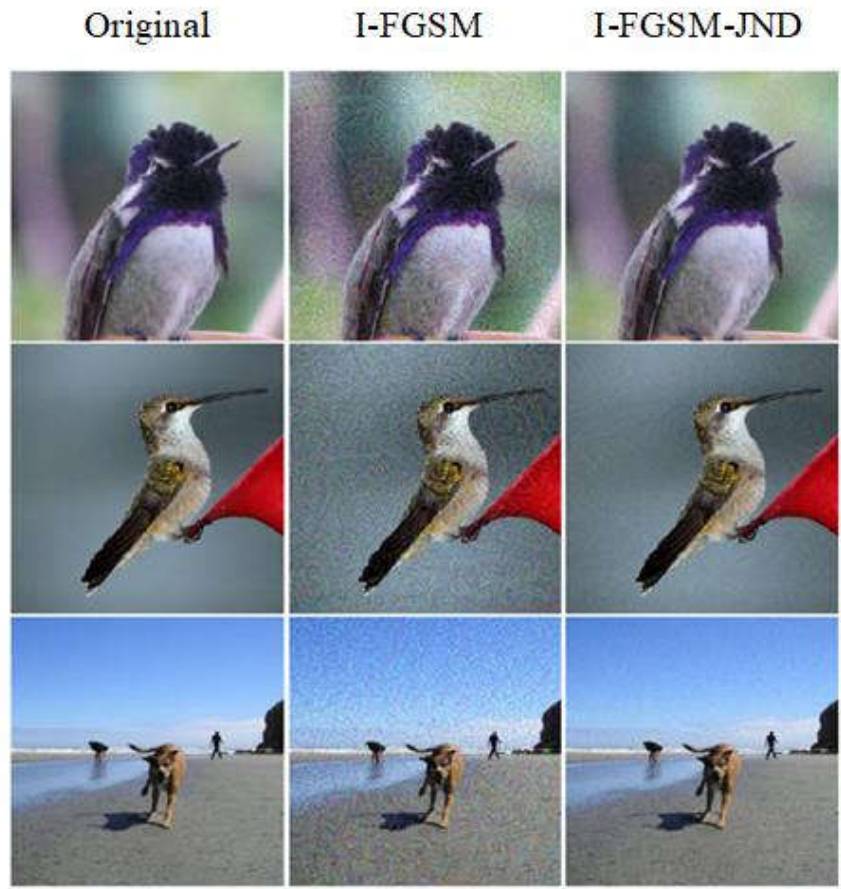

Figure 3. I-FGSM vs. I-FGSM-JND on the MiniImageNet dataset.

The I-FGSM-JND is obtained by integrating JND coefficients into the I-FGSM. In Figure 3, the I-FGSM generates more obvious perturbations, especially in the smooth background region. However, the perturbations generated by the I-FGSM-JND primarily focus on regions of complex texture in the images (e.g., the "bird" in row 1), which is not sensitive to the HVS, and perturbations in it cannot be detected easily. And even in smooth regions like the body of the "bird", our I-FGSM-JND generates smaller and fewer perturbations in such regions.

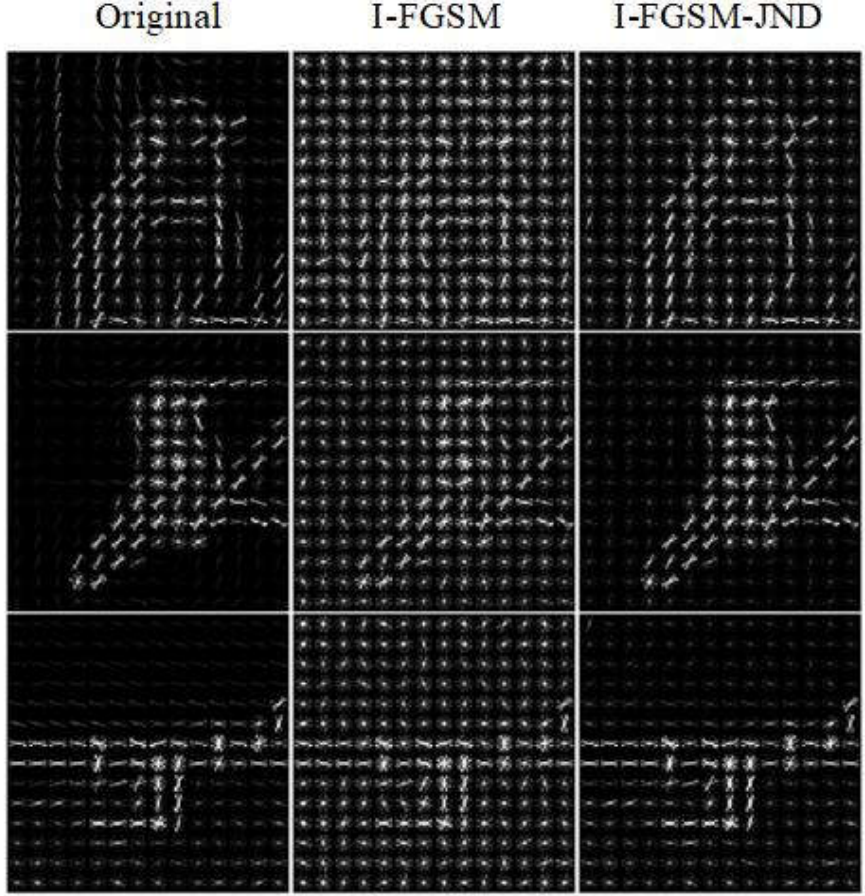

Figure 4. Histograms of oriented gradients generated by the original inputs, I-FGSM, and I-FGSM-JND in Figure 3.

From a different perspective, we can explain this phenomenon with the histograms of oriented gradients (HOG) [37], which is a feature descriptor of an image and reflects outline and texture information of an image. We herein configure the HOG basic settings with 8 orientations, 16×16 pixels per cell and 1×1 cell per block. In Figure 4, even though the HOG of the adversarial examples (e.g., still the "bird" in row 1) generated by the I-FGSM-JND can mainly be perturbed by a small noise texture in the background, the outline of "bird" can be recognized. By contrast, FGSM-JND's adversarial examples are covered with noise but cannot be recognized, that is, all the magnitudes and directions of textures are messy and even we can't distinguish the target and background. On the other hand, the HOG descriptors of the "bird" in row 2 and the "dog" in row 3 are clearer than that of the "bird" in row 1, especially in the background regions. It is most likely that the background in row 1 is more complex, where JND coefficients is larger and we can add more noise. The texture complexity reflects the information of the edge, which is related with gradient. Thus, the gradients distribution of adversarial examples generated by the I-FGSM-JND is more similar to those of the image inputs.

### 3.1.3 Original Methods vs. Improvement Methods

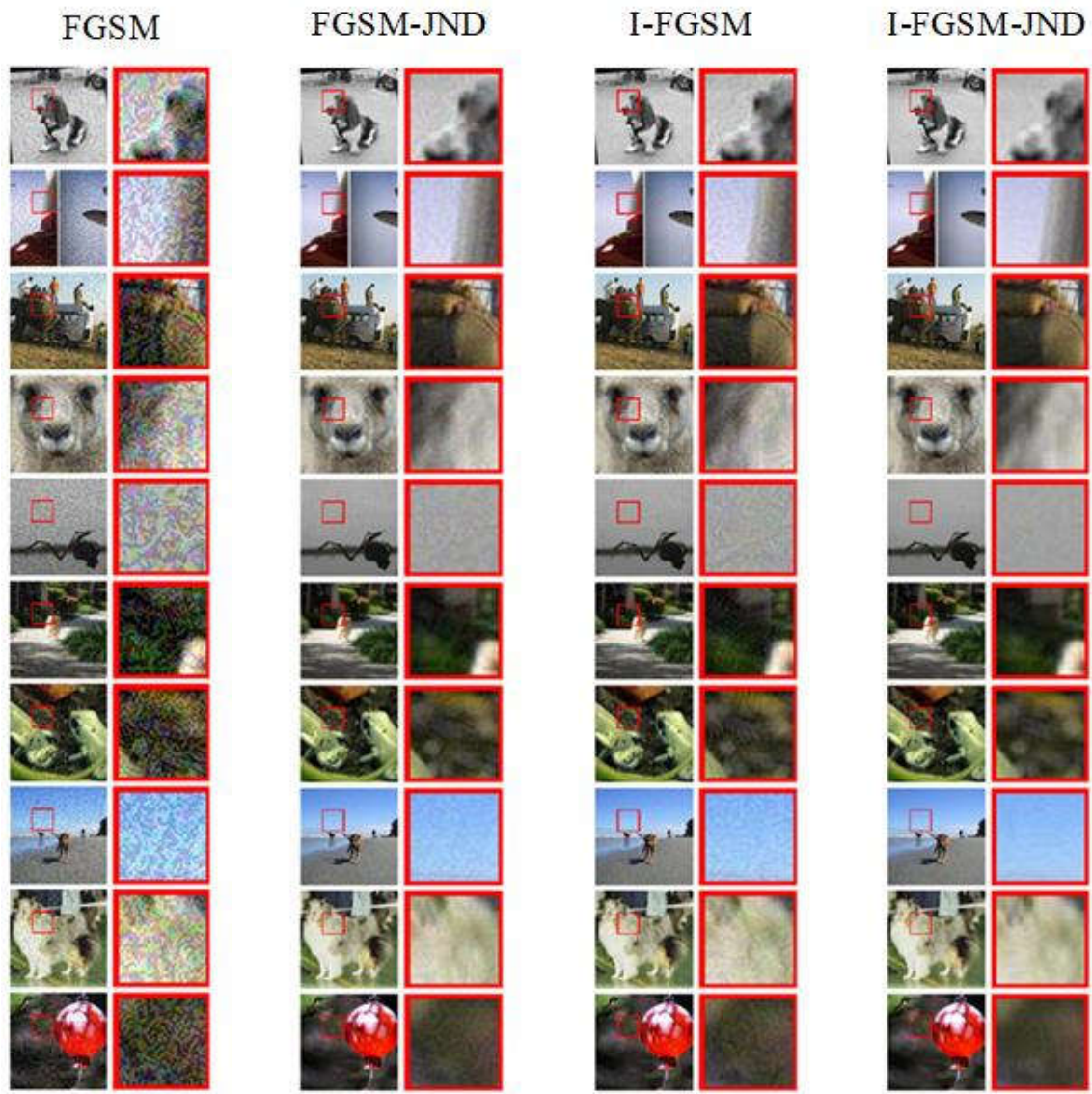

Figure 5. Ten adversarial examples were generated by the FGSM, FGSM-JND, I-FGSM, and I-FGSM-JND with epsilon 0.1; their local enlarged images on the MiniImageNet dataset are shown on the right orderly.

In Figure 5, we select 10 adversarial examples randomly and enlarge their local regions (marked by a red box in the same place) to see more information in detail. For example, in row 8, we enlarge the sky to observe. The FGSM method generates distinct perturbations and the I-FGSM can produce more refined perturbations by iterating the FGSM method, which also proves that it is useful to iterate. To our surprise, integrating JND coefficients into the constraint, we can get smaller perturbations than the I-FGSM method. For all images, we can conclude that our AdvJND algorithm improves the image quality obviously, especially in smooth regions with simple texture. And I-FGSM-JND algorithm performs best. There is no doubt that it works when we take the JND coefficients as a priori information to control the distribution of gradients.

Table 1. Comparison of recognition accuracy between the original attack and AdvJND attack on the FashionMNIST, CIFAR 10, and MiniImageNet datasets.

| Attack Methods | Epsilon | FashionMNIST/ LeNet5 | CIFAR10/ VGG16 | MiniImageNet/ Inception_v3 |
|---|---|---|---|---|
| Non-attack | 0.0 | 92.33 | 83.4 | 97.82 |
| FGSM | 0.2 | 12.94 | 9.02 | 43.64 |
| FGSM-JND |  | 29.48 | 9.22 | 58.49 |
| I-FGSM |  | 5.69 | 7.51 | 1.3 |
| I-FGSM-JND |  | 16.57 | 7.52 | 2.44 |

As shown in Table 1, the non-attack method means taking the original images as inputs without epsilon, and the attack success rate, namely (1-recognition accuracy), of the AdvJND algorithm is lower than or equivalent to that of the original attack method, which sacrifices a little attack success rates to improve the images fidelity. This is especially obvious in the FGSM and FGSM-JND. Because the FGSM is a one-step attack method, its effect on the attack success rate is larger than that on the image fidelity, which leads the gap of the attack success rate between the FGSM and FGSM-JND a little large. And by iterating, the attack success rate is higher and the image fidelity becomes better, meanwhile, the gap of the attack success rate between the I-FGSM and I-FGSM-JND decreases.

On the other hand, the performance of the FashionMNIST dataset, whether the attack success rate or the gap of the attack success rate between the original attack algorithm and our AdvJND algorithm, is worse than other datasets. It can be considered that the improvement effect of our AdvJND algorithm is a little critical about images because the JND coefficients are related to the the texture complexity of the image. However, such FashionMNIST dataset prefers simple textures and smooth backgrounds, and the MiniImageNet dataset includes more practical images in our real life with more complex textures. We know that the function of the JND coefficients are small in smooth images and the effects of the JND coefficients are not obvious, which explains why our AdvJND algorithm performs better on the MiniImageNet and CIFAR10 datasets than the FashionMNIST dataset.

### 3.1.4 Epsilon in AdvJND

The epsilon is crucial for improving the attack success rate. In this section, we present the attack success rate and image fidelity of AdvJND attacks by changing the epsilon value.

When the epsilon increases from 0.01 to 0.2, the attack success rate improves, too. Simultaneously, the gap between the I-FGSM and I-FGSM-JND decreases gradually. When the epsilon is 0.2, the difference in the attack success rate between the I-FGSM and I-FGSM-JND is less than 0.009. However, in terms of image quality (in Figure 6), the adversarial examples generated by the I-FGSM with epsilon 0.01 and those by the I-FGSM-JND with epsilon 0.2 with higher attack success rate are similar.

Therefore, adversarial examples generated by AdvJND are more similar to the original inputs when the attack success rates of the original attack and AdvJND attack are equivalent. In other words, by embedding the a priori information of the human eyes' subjective feeling, the algorithm based on AdvJND attack is more effective for alleviating the tradeoff between the attack success rate and image fidelity and achieves to generate adversarial examples with more higher image quality.

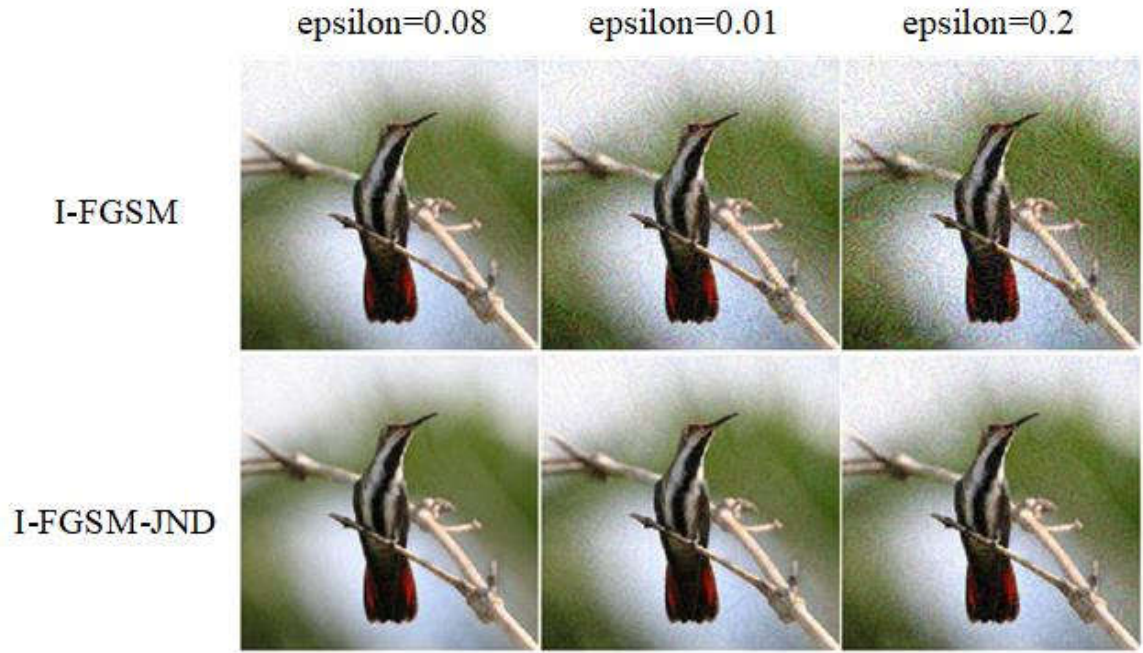

Figure 6. Adversarial examples generated by the I-FGSM and I-FGSM-JND with epsilon 0.01, 0.08, and 0.2

### 3.2 I-FGSM-JND vs. DeepFool

In Figure 7, the adversarial examples generated by the I-FGSM-JND (with epsilon 0.08) and DeepFool are similar. The attack success rate of the I-FGSM-JND algorithm is slightly higher than that of DeepFool, but the average of time consuming for the I-FGSM-JND algorithm to generate an adversarial example is approximately only half that of the DeepFool (in Table 2). The times are computed using a Nvidia GTX 1080Ti GPU. This is because DeepFool takes the smallest distance to the nearest classification boundary as the minimum perturbations. So it must traverse the classification boundary and obtain the smallest distance. In case of the situation of 1000 classes, the disadvantage of time-consuming will be more obvious. Thus, the efficiency of the I-FGSM-JND algorithm is significantly higher than that of DeepFool, and the I-FGSM-JND is more suitable as a universal attack method.

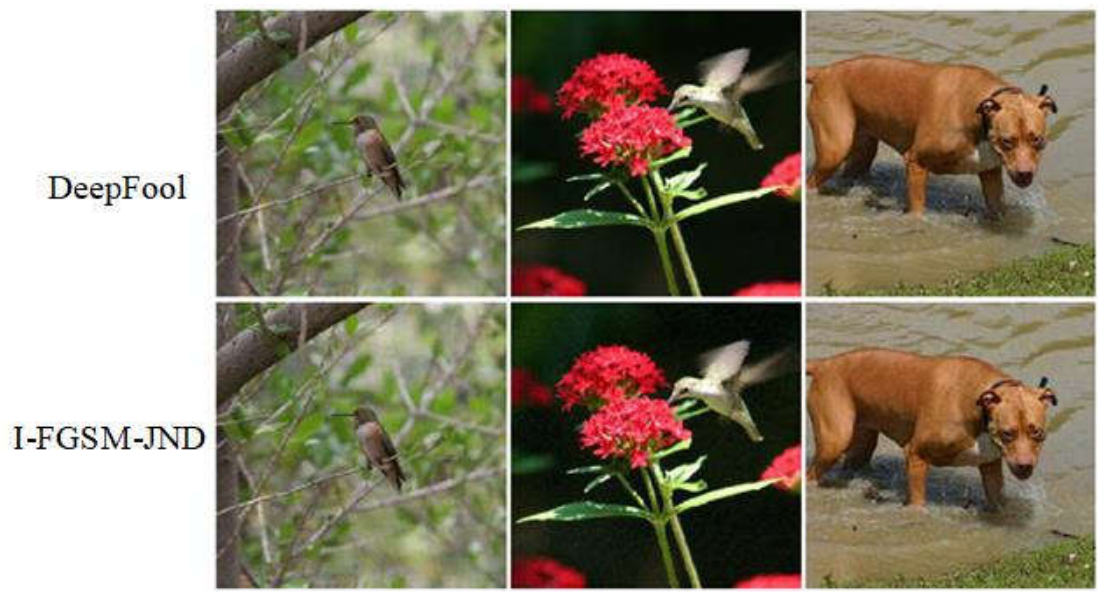

Figure 7. I-FGSM-JND vs. DeepFool on the MiniImageNet dataset.

Table 2. The efficiency of generating adversarial examples with the I-FGSM and DeepFool.

| Method | Attack Success Rate (%) | Average Time of Generating an Adversarial Example(s) |
|---|---|---|
| I-FGSM-JND | 97.45 | 0.7 |
| DeepFool | 96.36 | 1.41 |

Similar to integrating the JND coefficients in the $L_\infty$ norm, the subjective visual information of the human eyes is used as a priori information to improve the image quality of the adversarial examples. Furthermore, we can consider to embed the appropriate visual model coefficients into the $L_2$ norm constraint as a priori information which can provide a better search strategy or reduce the search space to decrease the iteration or traversal times to improve the efficiency.

## 4. Conclusions

Large perturbations lead the adversarial examples' high attack success rate and bad image fidelity with poor concealment. To alleviate the tradeoff between the attack success rate and image fidelity, we herein proposed an adversarial attack method using AdvJND and used JND coefficients to relate the subjective feeling of human eyes and the image quality evaluation metric. The human eyes are not sensitive to changes in complex texture regions, which provides a chance for us to embed more noise in these regions. Our experimental results demonstrated that the HOG descriptors of adversarial examples generated by the AdvJND algorithm were similar to those of the original inputs; thus, noise could be hidden effectively in the original inputs. Our approach can be incorporated into the new proposed $L_\infty$ norm-based attack method to build adversarial examples that are similar to the original inputs. In future work, other metrics of human visual evaluation can be integrated into the $L_2$ norm constraint to improve the efficiency of generating adversarial examples.

## Data Availability

The information about the FashionMNIST, CIFAR10 and ImageNet datasets are available and can be downloaded from https://github.com/zalandoresearch/fashion-mnist, https://www.cs.toronto.edu/~kriz/cifar.html, http://www.imagenet.org/download.

## Conflicts of Interest

The authors declare that there is no conflict of interest regarding the publication of this paper.

## Funding Statement

This work was supported by the National Key Research and Development Plan of China (No. 2017YFB1002502), and the Natural Science Foundation of Henan Province of China (No.162300410333).